\documentclass[runningheads]{llncs}
\usepackage{graphicx}
\usepackage{amsmath,amssymb} 
\usepackage{color}
\usepackage[width=122mm,left=12mm,paperwidth=146mm,height=193mm,top=12mm,paperheight=217mm]{geometry}

\usepackage{algorithm}
\usepackage{algpseudocode}
\algrenewcommand\algorithmicindent{.9em}%

\usepackage{comment}
\usepackage[british,UKenglish,USenglish,english,american]{babel}
\usepackage{multirow}

\newcommand{\tabincell}[2]{\begin{tabular}{@{}#1@{}}#2\end{tabular}}

\newlength\savewidth\newcommand\shline{\noalign{\global\savewidth\arrayrulewidth
  \global\arrayrulewidth 1pt}\hline\noalign{\global\arrayrulewidth\savewidth}}

\newcommand{\ie}{{\itshape i.e.}}
\newcommand{\eg}{{\itshape e.g.}}
\begin{document}
\pagestyle{headings}
\mainmatter

\title{Learning Region Features for Object Detection} 

\titlerunning{Learning Region Features for Object Detection}

\authorrunning{Jiayuan Gu, Han Hu, Liwei Wang, Yichen Wei and Jifeng Dai}

\author{Jiayuan Gu$^{1}$\thanks{This work is done when Jiayuan Gu is an intern at Microsoft Research Asia.}, Han Hu$^2$, Liwei Wang$^1$, Yichen Wei$^2$ and Jifeng Dai$^2$ }

\institute{$^1$Key Laboratory of Machine Perception, School of EECS, Peking University\\
	\email{ \{gujiayuan,wanglw\}@pku.edu.cn} \\
$^2$Microsoft Research Asia\\
	\email{ \{hanhu, yichenw, jifdai\}@microsoft.com}
}


\maketitle

\begin{abstract}
While most steps in the modern object detection methods are learnable, the region feature extraction step remains largely hand-crafted, featured by RoI pooling methods. This work proposes a general viewpoint that unifies existing region feature extraction methods and a novel method that is end-to-end learnable. The proposed method removes most heuristic choices and outperforms its RoI pooling counterparts. It moves further towards \emph{fully learnable object detection}.
\end{abstract}

\section{Introduction}
A noteworthy trait in the deep learning era is that many hand-crafted features, algorithm components, and design choices, are replaced by their data-driven and learnable counterparts. The evolution of object detection is a good example. Currently, the leading region-based object detection paradigm~\cite{girshick2014rich,he2014spatial,girshick2015fast,ren2015faster,dai2016rfcn,lin2016feature,dai2017deformable,he2017mask} consists of five steps, namely, image feature generation, region proposal generation, region feature extraction, region recognition, and duplicate removal. Most steps become learnable in recent years, including image feature generation~\cite{girshick2015fast}, region proposal~\cite{szegedy2014scalable,erhan2014scalable,ren2015faster}, and duplicate removal~\cite{hosang2017learning,hu2018relation}. Note that region recognition step is learning based in nature.

The region feature extraction step remains largely hand-crafted. The current practice, RoI (regions of interest) pooling~\cite{girshick2015fast}, as well as its variants~\cite{he2014spatial,he2017mask}, divides a region into regular grid bins, computes features of the bin from the image features located nearby to the bin via heuristic rules (avg, max, bilinear interpolation~\cite{he2017mask,dai2017deformable}, etc), and concatenates such features from all the bins as the region features. The process is intuitive and works well, but is more like rules of thumb. There is no clear evidence that it is optimal in some sensible way.

The recent work of deformable RoI pooling~\cite{dai2017deformable} introduces a bin-wise offset that is adaptively learnt from the image content. The approach is shown better than its RoI pooling counterpart. It reveals the potential of making the region feature extraction step \emph{learnable}. However, its form  still resembles the regular grid based pooling. The learnable part is limited to bin offsets only.

This work studies \emph{fully learnable} region feature extraction. It aims to improve the performance and enhance the understanding of this step. It makes the following two contributions.

First, a general viewpoint on region feature extraction is proposed. The feature of each bin (or in a general sense, part) of the region is formulated as a weighted summation of image features on different positions over the whole image. Most (if not all) previous region feature extraction methods are shown to be specialization of this formulation by specifying the weights in different ways, mostly hand-crafted.

Based on the viewpoint, the second contribution is a learnable module that represents the weights in terms of the RoI and image features. The weights are affected by two factors: the geometric relation between the RoI and image positions, as well as the image features themselves. The first is modeled using an attention model as motivated by~\cite{vaswani2017attention,hu2018relation}. The second is exploited by simply adding one convolution layer over the input image features, as motivated by~\cite{dai2017deformable}.

The proposed method removes most heuristic choices in the previous RoI pooling methods and moves further towards \emph{fully learnable object detection}. Extensive experiments show that it outperforms its RoI pooling counterparts. While a naive implementation is computationally expensive, an efficient sparse sampling implementation is proposed with little degradation in accuracy. Moreover, qualitative and quantitative analysis on the learnt weights shows that it is feasible and effective to learn the spatial distribution of such weights from data, instead of designing them manually.

\section{A General Viewpoint on Region Feature Extraction}
\label{sec.general_viewpoint}

Image feature generation step outputs feature maps $\mathbf{x}$ of spatial size $H\times W$ (usually $16\times$ smaller than that of the original image due to down sampling of the network~\cite{ren2015faster}) and $C_f$ channels. Region proposal generation step finds a number of regions of interest (RoI), each a four dimensional bounding box $b$.

In general, the region feature extraction step generates features $\mathbf{y}(b)$ from $\mathbf{x}$ and an RoI $b$ as

\begin{equation}
\mathbf{y}(b) =  \text{RegionFeat}(\mathbf{x}, b).
\end{equation}

Typically, $\mathbf{y}(b)$ is of dimension $K\times C_f$. The channel number is kept the same as $C_f$ in $\mathbf{x}$ and $K$ represents the number of \emph{spatial parts} of the region. Each part feature $\mathbf{y}_k(b)$ is a partial observation of the region. For example, $K$ is the number of bins (\eg, $7\times 7$) in the current RoI pooling practice. Each part is a bin in the regular grid of the RoI. Each $\mathbf{y}_k(b)$ is generated from image features in $\mathbf{x}$ within the bin.

The concepts above can be generalized. A part does not need to have a regular shape. The part feature $\mathbf{y}_k(b)$ does not need to come from certain spatial positions in $\mathbf{x}$. Even, the union of all the parts does not need to be the RoI itself. A general formulation is to treat the part feature as the weighted summation of image features $\mathbf{x}$ over all positions within a support region $\Omega_b$, as
\begin{equation}
\label{eq.weighted_average_part_feature}
\mathbf{y}_k(b) = \sum_{p \in \Omega_b}w_k(b,p,\mathbf{x})\odot \mathbf{x}(p).
\end{equation}

Here, $\Omega_b$ is the supporting region. It could simply be the RoI itself or include more context, even the entire image. $p$ enumerates the spatial positions within $\Omega_b$. $w_k(b,p,x)$ is the weight to sum the image feature $\mathbf{x}(p)$ at the position $p$. $\odot$ denotes element-wise multiplication. Note that the weights are assumed normalized, \ie, $\sum_{p \in \Omega_b}w_k(b,p,x) = 1$.

We show that various RoI pooling methods~\cite{girshick2015fast,he2014spatial,he2017mask,dai2017deformable} are specializations of Eq.~(\ref{eq.weighted_average_part_feature}). The supporting region $\Omega_b$ and the weight $w_k(\cdot)$ are realized differently in these methods, mostly in hand-crafted ways.

\paragraph{Regular RoI Pooling~\cite{girshick2015fast}} The supporting region $\Omega_b$ is the RoI itself. It is divided into regular grid bins (\eg, $7 \times 7$). Each part feature $\mathbf{y}_k(b)$ is computed as max or average of all image features $\mathbf{x}(p)$ where $p$ is within the $k^{th}$ bin.

Taking averaging pooling as an example, the weight in Eq.~(\ref{eq.weighted_average_part_feature}) is

\begin{equation}
w_k(b, p) =
\begin{cases}
1/|R_{bk}| & \text{ if $p \in R_{bk}$} \\
0 & \text{else}
\end{cases}
\label{eq.weight_in_roi_pooling}
\end{equation}

Here, $R_{bk}$ is the set of all positions within the $k^{th}$ bin of the grid.

The regular pooling is flawed in that it cannot distinguish between very close RoIs due to spatial down sampling in the networks, \ie, the spatial resolution of the image feature $\mathbf{x}$ is usually smaller (\eg, $16\times$) than that of the original image. If two RoIs' distance is smaller than $16$ pixels, their $R_{bk}$s are the same, and so are their features.

\paragraph{Spatial Pyramid Pooling~\cite{he2014spatial}} Because it simply applies the regular RoI pooling on different levels of grid divisions, it can be expressed via simple modification of Eq.~(\ref{eq.weighted_average_part_feature}) and~(\ref{eq.weight_in_roi_pooling}). Details are irrelevant and omitted here.

\paragraph{Aligned RoI Pooling~\cite{he2017mask}} It remedies the quantization issue in the regular RoI pooling above by bilinear interpolation at fractionally sampled positions within each $R_{bk}$. For simplicity, we assume that each bin only samples one point, \ie, its center $(u_{bk}, v_{bk})$\footnote{In practical implementation~\cite{he2017mask}, multiple (\eg, 4) points are sampled within each bin and their features are averaged as the bin feature. This is beneficial as more image position features get back-propagated gradients.}. Let the position $p=(u_p,v_p)$. The weight in Eq.~(\ref{eq.weighted_average_part_feature}) is

\begin{equation}
w_k(b, p) = g(u_p, u_{bk}) \cdot g(v_p, v_{bk}),
\label{eq.weight_in_roi_aligned_pooling}
\end{equation}

where $g(a,b) = \max(0, 1-|a-b|)$ denotes the 1-D bilinear interpolation weight. Note that the weight in Eq.~(\ref{eq.weight_in_roi_aligned_pooling}) is only non-zero for the four positions immediately surrounding the sampling point $(u_{bk}, v_{bk})$.

Because the weight in Eq.~(\ref{eq.weight_in_roi_aligned_pooling}) depends on the bin center $(u_{bk}, v_{bk})$, the region features are sensitive to even subtle changes in the position of the RoI. Thus, aligned pooling outperforms its regular pooling counterpart~\cite{he2017mask}.

Note that everything till now is hand-crafted. Also, image feature $\mathbf{x}$ is not used in $w_k(\cdot)$ in Eq.~(\ref{eq.weight_in_roi_pooling}) and~(\ref{eq.weight_in_roi_aligned_pooling}).

\paragraph{Deformable RoI pooling~\cite{dai2017deformable}} It generalizes aligned RoI pooling by learning an offset $(\delta u_{bk}, \delta v_{bk})$ for each bin and adding it to the bin center. The weight in Eq.~(\ref{eq.weight_in_roi_aligned_pooling}) is extended to

\begin{equation}
w_k(b,p,\mathbf{x}) = g(u_p, u_{bk} + \delta u_{bk}) \cdot g(v_p, v_{bk} + \delta v_{bk}).
\end{equation}

The image feature $\mathbf{x}$ appears here because the offsets are produced by a learnable submodule applied on the image feature $\mathbf{x}$. Specifically, the submodule starts with a regular RoI pooling to extract an initial region feature from image feature, which is then used to regress offsets through an additional learnable fully connected (fc) layer.

As the weight and the offsets depend on the image features now and they are learnt end-to-end, object shape deformation is better modeled, adaptively according to the image content. It is shown that deformable RoI pooling outperforms its aligned version~\cite{dai2017deformable}. Note that when the offset learning rate is zero, deformable RoI pooling strictly degenerates to aligned RoI pooling.

Also note that the supporting region $\Omega_b$ is no longer the RoI as in regular and aligned pooling, but potentially spans the whole image, because the learnt offsets could be arbitrarily large, in principle.

\subsection{More Related Works}
Besides the RoI pooling methods reviewed above, there are more region feature extraction methods that can be thought of specializations of Eq.~(\ref{eq.weighted_average_part_feature}) or its more general extension.

\paragraph{Region Feature Extraction in One-stage Object Detection~\cite{liu2016ssd,redmon2016you,lin2017focal}} As opposed to the two-stage or region based object detection paradigm, another paradigm is one-stage or dense sliding window based. Because the number of windows (regions) is huge, each region feature is simply set as the image feature on the region's center point, which can be specialized from Eq.~(\ref{eq.weighted_average_part_feature}) as $K=1$, $\Omega_b = \{\text{center}(b)\}$. This is much faster but less accurate than RoI pooling methods.

\paragraph{Pooling using Non-grid Bins~\cite{chen2017masklab,wu2017interpretable}} These methods are similar to regular pooling but change the definition of $R_{bk}$ in Eq.~(\ref{eq.weight_in_roi_pooling}) to be non-grid. For example, MaskLab~\cite{chen2017masklab} uses triangle-shaped bins other than rectangle ones. It shows better balance in encoding center-close and center-distant subregions. In Interpretable R-CNN~\cite{wu2017interpretable}, the non-grid bins are generated from the grammar defined by an AND-OR graph model.

\paragraph{MNC~\cite{dai2016mnc}} It is similar as regular RoI pooling.  The difference is that only the bins inside the mask use Eq.~(\ref{eq.weight_in_roi_pooling}) to compute weights. The weights of the bins outside are zeros. This equals to relax the normalization assumption on $w_k$.

\paragraph{Position Sensitive RoI Pooling~\cite{dai2016rfcn,li2017light}} It is similar as regular RoI pooling. The difference is that each bin only corresponds to a subset of channels in the image feature $\mathbf{x}$, instead of all channels. This can be expressed by extending Eq.~(\ref{eq.weighted_average_part_feature}) as
\begin{equation}
\label{eq.weighted_average_part_feature_extension}
\mathbf{y}_k(b) = \sum_{p \in \Omega_b}w_k(b,p,\mathbf{x}_k)\odot \mathbf{x}_k(p),
\end{equation}
where $\mathbf{x}_k$ only contains a subset of channels in $\mathbf{x}$, according to the $k^{\text{th}}$ bin.

\begin{figure}[t]
\centering
\includegraphics[width=0.98\textwidth]{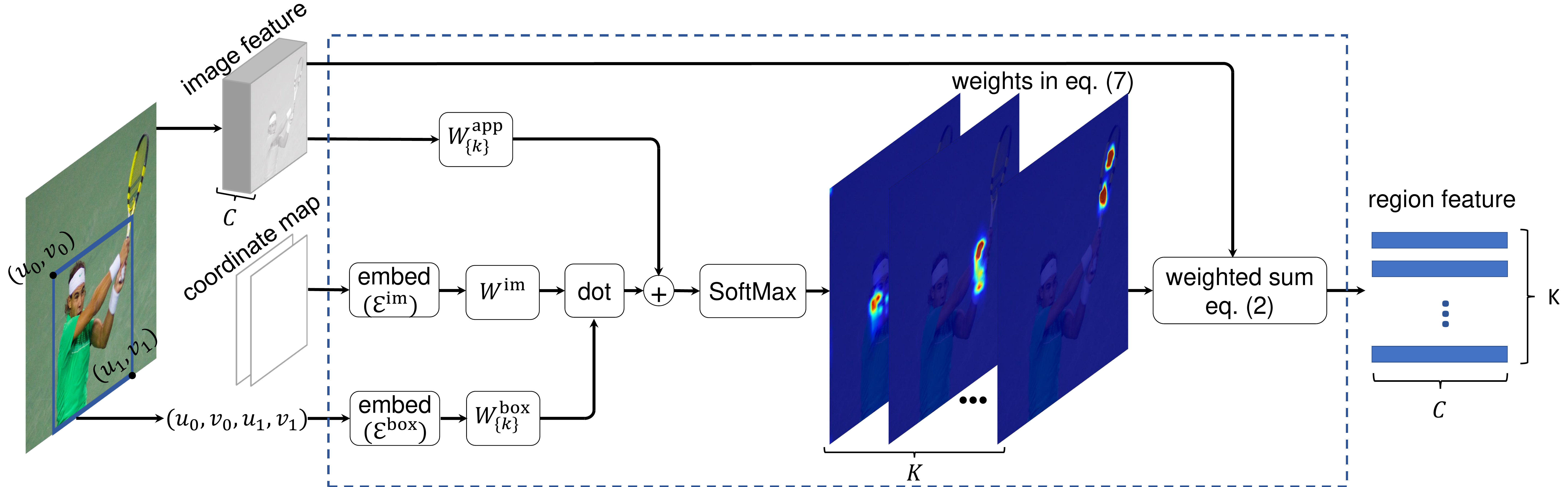}
\caption{Illustration of the proposed region feature extraction module in Eq.~(\ref{eq.weighted_average_part_feature}) and~(\ref{eq.att_weight}).}
\label{fig.algorithm_illustration}
\end{figure}

\section{Learning Region Features}
\label{sec.learning_region_features}

Regular and aligned RoI pooling are fully hand-crafted. Deformable RoI pooling introduces a learnable component, but its form is still largely limited by the regular grid. In this work, we seek to learn the weight $w_k(b,p,\mathbf{x})$ in Eq.~(\ref{eq.weighted_average_part_feature}) with minimum hand crafting.

Intuitively, we consider two factors that should affect the weight. First, the geometric relation between the position $p$ and RoI box $b$ is certainly critical. For example, positions within $b$ should contribute more than those far away from it. Second, the image feature $\mathbf{x}$ should be adaptively used. This is motivated by the effectiveness of deformable RoI pooling~\cite{dai2017deformable}.

Therefore, the weight is modeled as the exponential of the sum of two terms

\begin{equation}
\label{eq.att_weight}
w_k(b,p,\mathbf{x}) \propto \exp(G_k(b,p)+A_k(\mathbf{x},p)).
\end{equation}

The first term $G_k(b,p)$ in Eq.~(\ref{eq.att_weight}) captures \emph{geometric relation} as

\begin{equation}
\label{eq.geo_att}
G_k(b,p) = \langle W^{\text{box}}_{k}\cdot \mathcal{E}^{\text{box}}(b), W^{\text{im}}\cdot \mathcal{E}^{\text{im}}(p) \rangle.
\end{equation}

There are three steps. First, the box and image positions are embedded into high dimensional spaces similarly as in~\cite{vaswani2017attention,hu2018relation}. The embedding is performed by applying sine and cosine functions of varying wavelengths to a scalar $z$, as

\begin{equation*}
\mathcal{E}_{2i}(z) = \sin \Large ( \frac{z}{1000^{2i/C_{\mathcal{E}}}} \Large ), \quad \mathcal{E}_{2i+1}(z) = \cos \Large (\frac{z}{1000^{2i/C_{\mathcal{E}}}} \Large ).
\end{equation*}

The embedding vector $\mathcal{E}(z)$ is of dimension $C_{\mathcal{E}}$. The subscript $i$ above ranges from $0$ to $C_{\mathcal{E}}/2-1$. The image position $p$ is embedded into a vector $\mathcal{E}^{\text{im}} (p)$ of dimension $2\cdot C_{\mathcal{E}}$, as $p$ has two coordinates. Similarly, each RoI box $b$ is embedded into a vector $\mathcal{E}^{\text{box}}(b)$ of dimension $4\cdot C_{\mathcal{E}}$.

Second, the embedding vectors $\mathcal{E}^{\text{im}}(p)$ and $\mathcal{E}^{\text{box}}(b)$ are linearly transformed by weight matrices $W^{\text{im}}$ and $W^{\text{box}}_k$, respectively, which are learnable. The transformed vectors are of the same dimension $C_g$. Note that the term $W^{\text{box}}_{k}\cdot \mathcal{E}^{\text{box}}(b)$ has high complexity because the $\mathcal{E}^{\text{box}}(b)$'s dimension $4\cdot C_{\mathcal{E}}$ is large. In our implementation, we decompose $W^{\text{box}}_k$ as $W^{\text{box}}_k = \hat{W}^{\text{box}}_k V^{\text{box}}$. Note that $V^{\text{box}}$ is shared for all the parts. It does not have subscript $k$. Its output dimension is set to $C_{\mathcal{E}}$. In this way, both computation and the amount of parameters are reduced for the term $W^{\text{box}}_{k}\cdot \mathcal{E}^{\text{box}}(b)$.

Last, the inner product of the two transformed vectors is treated as the geometric relation weight.

Eq.~(\ref{eq.geo_att}) is basically an attention model~\cite{vaswani2017attention,hu2018relation}, which is a good tool to capture dependency between distant or heterogeneous elements, e.g., words from different languages~\cite{vaswani2017attention}, RoIs with variable locations/sizes/aspect ratios~\cite{hu2018relation}, and etc, and hence naturally bridges the target of building connections between 4D bounding box coordinates and 2D image positions in our problem. Extensive experiments show that the geometric relations between RoIs and image positions are well captured by the attention model.

The second term $A_k(\mathbf{x},p)$ in Eq.~(\ref{eq.att_weight}) uses the \emph{image features} adaptively. It applies an $1\times 1$ convolution on the image feature,

\begin{equation}
\label{eq.app_att}
A_k(\mathbf{x},p) = W^{\text{app}}_k \cdot \mathbf{x}(p),
\end{equation}
where $ W^{\text{app}}_k$ denotes the convolution kernel weights, which are learnable.

The proposed region feature extraction module is illustrated in Figure~\ref{fig.algorithm_illustration}. During training, the image features $\mathbf{x}$ and the parameters in the module ($W^{\text{box}}_k$, $W^{\text{im}}$, and $W^{\text{app}}_k$) are updated simultaneously.

\subsection{Complexity Analysis and an Efficient Implementation}
\label{sec.compute_complexity}

\renewcommand{\arraystretch}{1.2}
\begin{table}[t]

\begin{center}
\scriptsize
\begin{tabular}{c|l|c || c|l|c}
\shline
\text{notation} & \text{description} & \tabincell{l}{typical\\values} & \text{notation} & \text{description} & \tabincell{l}{typical\\values}\\
\shline
$|\Omega_b|$ & size of support region & hundreds &
$N$ & $\#$RoIs & 300 \\
$H$ & height of image feature $\mathbf{x}$ & dozens &
$K$ & $\#$parts/bins & 49  \\
$W$ & width of image feature $\mathbf{x}$ & dozens &
$C_{\mathcal{E}}$ & embed dim. in Eq.~(\ref{eq.geo_att}) & 512 \\
$C_f$ & $\#$channels of image feature $\mathbf{x}$ & 256 &
$C_g$ & transform dim. in Eq.~(\ref{eq.geo_att}) & 256 \\
\shline
\end{tabular}
\vspace{.5em}

\small
\begin{tabular}{l|c|c|c}
\shline
\text{~~~~~~~~~~~~~~~~~~~~module} & \tabincell{c}{computational\\complexity} & \tabincell{c}{\textit{naive}\\($|\Omega_b|$=HW)} & \tabincell{c}{\textit{efficient}\\($|\Omega_b|$=200$^\dag$)} \\
\shline
\hline
(P1) { \scriptsize transform position embedding in Eq.~(\ref{eq.geo_att})}  & $2 H W C_{\mathcal{E}} C_g$ & 0.59G & 0.59G\\
(P2) {\scriptsize transform RoI box embedding in Eq.~(\ref{eq.geo_att})} & \scriptsize $N C_{\mathcal{E}} (KC_g+4C_{\mathcal{E}})$* & 2.1G & 2.1G\\
(P3) inner product in Eq.~(\ref{eq.geo_att}) &  $N K |\Omega_b| C_g$ & 7.2G & \textbf{0.72G} \\
(P4) appearance usage in Eq.~(\ref{eq.app_att}) & $H W K C_f$ &0.03G & 0.03G\\
(P5) weighted aggregation in Eq.~(\ref{eq.weighted_average_part_feature}) &  $N K |\Omega_b| C_f$ & 7.2G & \textbf{0.72G}\\
\hline
~~~~~~~~~~~~~~~~~~~~~~sum & & 17.1G & 4.16G \\
\hline
\end{tabular}
\end{center}
\caption{\textbf{Top}: description and typical values of main variables. \textbf{Bottom}: computational complexity of the proposed method. $\dag$Using default maximum sample numbers as in Eq.~(\ref{eq.stride_b_in}) and (\ref{eq.stride_b_out}), the average actual sample number is about 200. See also Table~\ref{table.aggregation_point_sampling}. *Note that we decompose $W^{\text{box}}_k$ as $W^{\text{box}}_k = \hat{W}^{\text{box}}_k V^{\text{box}}$, and the total computational cost is the sum of two matrix multiplications $V^{\text{box}} \cdot \mathcal{E}^{\text{box}}$ (the multiplication result is denoted as $\hat{\mathcal{E}}^{\text{box}}$) and $\hat{W}^{\text{box}}_k \cdot \hat{\mathcal{E}}^{\text{box}}$. See also Section~\ref{sec.learning_region_features} for details.}
\label{table.complexity}
\end{table}

The computational complexity of the proposed region feature extraction module is summarized in Table~\ref{table.complexity}. Note that $A_k(x,p)$ and $W^{\text{im}}\cdot \mathcal{E}^{\text{im}}(p)$ are computed over all the positions in the image feature $\mathbf{x}$ and shared for all RoIs.

A \emph{naive} implementation needs to enumerate all the positions in $\Omega_b$. When $\Omega_b$ spans the whole image feature $\mathbf{x}$ densely, its size is $H\times W$ and typically a few thousands. This incurs heavy computational overhead for step 3 and 5 in Table~\ref{table.complexity}. An \emph{efficient} implementation is to sparsely sample the positions in $\Omega_b$, during the looping of $p$ in Eq.~(\ref{eq.weighted_average_part_feature}). Intuitively, the sampling points within the RoI should be denser and those outside could be sparser. Thus, $\Omega_b$ is split into two sets as $\Omega_b=\Omega_b^{\text{In}}\cup \Omega_b^{\text{Out}}$, which contain the positions within and outside of the RoI, respectively. Note that $\Omega_b^{\text{Out}}$ represents the context of the RoI. It could be either empty when $\Omega_b$ is the RoI or span the entire image when $\Omega_b$ does, too.

Complexity is controlled by specifying a maximum number of sampling positions for $\Omega_b^{\text{In}}$ and $\Omega_b^{\text{Out}}$, respectively (by default, 196 for both). Given an RoI $b$, the positions in $\Omega_b^{\text{In}}$ are sampled at stride values $stride_{\text{x}}^{b}$ and $stride_{\text{y}}^{b}$, in x and y directions, respectively. The stride values are determined as
\begin{equation}
\label{eq.stride_b_in}
stride_{\text{x}}^{b} = \lceil W_b/\sqrt{196} \rceil \text{ AND } stride_{\text{y}}^{b} = \lceil H_b/\sqrt{196} \rceil,
\end{equation}
where $W_b$ and $H_b$ are the width and height of the RoI. The sampling of $\Omega_b^{\text{Out}}$ is similar. Let $stride^{\text{out}}$ be the stride value, it is derived by,
\begin{equation}
\label{eq.stride_b_out}
stride^{\text{out}} = \lceil \sqrt{HW/196} \rceil.
\end{equation}

The sparse sampling of $\Omega_b$ effectively reduces the computational overhead. Especially, notice that many RoIs have smaller area than the maximum sampling number specified above. So the actual number of sampled positions of $\Omega_b^{\text{In}}$ in those RoIs is equal to their area, thus even smaller.

Experiments show that the accuracy of sparse sampling is very close to the naive dense sampling (see Table~\ref{table.aggregation_point_sampling}).

\section{Experiments}
All experiments are performed on COCO detection datasets~\cite{lin2014coco}. We follow the COCO 2017 dataset split: 115k images in the \textit{train} split for training; 5k images in the \textit{minival} split for validation; and 20k images in the \textit{test-dev} split for testing. In most experiments, we report the accuracy on the \textit{minival} split.

State-of-the-art Faster R-CNN~\cite{ren2015faster} and FPN~\cite{lin2016feature} object detectors are used. ResNet-50 and ResNet-101~\cite{he2016deep} are used as the backbone image feature extractor. By default, Faster R-CNN with ResNet-50 is utilized in ablation study.

For Faster R-CNN, following the practice in~\cite{dai2016rfcn,dai2017deformable}, the \textit{conv4} and \textit{conv5} image features are utilized for region proposal generation and object detection, respectively. The RPN branch is the same as in~\cite{ren2015faster,dai2016rfcn,dai2017deformable}. For object detection, the effective feature stride of \textit{conv5} is reduced from 32 pixels to 16 pixels. Specifically, at the beginning of the \textit{conv5} block, stride is changed from 2 to 1. The dilation of the convolutional filters in the \textit{conv5} block is changed from 1 to 2. On top of the \textit{conv5} feature maps, a randomly initialized $1 \times 1$ convolutional layer is added to reduce the dimension to 256-D. The proposed module is applied on top to extract regional features, where 49 bins are utilized by default. Two fully-connected (fc) layers of 1024-D, followed by the classification and the bounding box regression branches, are utilized as the detection head. The images are resized to 600 pixels at the shorter side if the longer side after resizing is less than or equal to 1000; otherwise resized to 1000 pixels at the longer side, in both training and inference~\cite{girshick2015fast}.

For FPN, a feature pyramid is built upon an input image of single resolution, by exploiting multi-scale feature maps generated by top-down and lateral connections. The RPN and Fast R-CNN heads are attached to the multi-scale feature maps, for proposing and detecting objects of varying sizes. Here we follow the network design in~\cite{lin2016feature}, and just replace RoI pooling by the proposed learnable region feature extraction module.

The images are resized to 800 pixels at the shorter side if the longer side after resizing is less than or equal to 1333; otherwise resized to 1333 pixels at the longer side, in both training and inference.

SGD training is performed on 4 GPUs with 1 image per GPU. Weight decay is $1 \times 10^{-4}$ and momentum is $0.9$. The added parameters in the learnable region feature extraction module, $W^{\text{box}}_k$, $W^{\text{im}}$, and $W_k^{\text{app}}$, are initialized by random Gaussian weights ($\sigma = 0.01$), and their learning rates are kept the same as the existing layers. In both Faster R-CNN and FPN, to facilitate experiments, separate networks are trained for region proposal generation and object detection, without sharing their features. In Faster R-CNN, 6 and 16 epochs are utilized to train the RPN and the object detection networks, respectively. The learning rates are set as $2\times 10^{-3}$ for the first $\frac{2}{3}$ iterations and $2\times 10^{-4}$ for the last $\frac{1}{3}$ iterations, for both region proposal and object detection networks. In FPN, 12 epochs are utilized to train both the RPN and the object detection networks, respectively. For both networks training, the learning rates start with $5\times 10^{-3}$ and decay twice at 8 and 10.667 epochs, respectively. Standard NMS with IoU threshold of 0.5 is utilized for duplication removal.

\renewcommand{\arraystretch}{1.1}
\begin{table}[t]
\small
\begin{center}
\begin{tabular}{l|c|cc|ccc}
 \tabincell{c}{method} & \tabincell{c}{\scriptsize mAP} &
\scriptsize mAP$_{50}$ & \scriptsize mAP$_{75}$ & \tabincell{c}{\scriptsize mAP$_S$} & \tabincell{c}{\scriptsize mAP$_M$} & \tabincell{c}{\scriptsize mAP$_L$} \\
\shline
\multicolumn{5}{l}{$1\times$ RoI} \\
\hline
regular RoI pooling & 29.8 & 52.2 & 29.9 & 10.4 & 32.6 & 47.8\\
aligned RoI pooling & 32.9 & 54.0 & 34.9 & 13.9 & 36.9 & 48.8\\
ours & \textbf{33.4} & 54.5 & 35.2 & 13.9 & 37.3 & 50.4\\
\shline
\multicolumn{5}{l}{$2\times$ RoI} \\
\hline
regular RoI pooling & 30.1 & 53.2 & 30.6 & 10.6 & 33.3 & 47.4\\
aligned RoI pooling & 32.8 & 54.6 & 35.1 & 14.2 & 37.0 & 48.5\\
ours & \textbf{33.8} & 55.1 & 35.8 & 14.2 & 37.8 & 51.1\\
\shline
\multicolumn{5}{l}{Whole image} \\
\hline
regular RoI pooling* & - & - & - & - & - & - \\
aligned RoI pooling* & - & - & - & - & - & -\\
ours & \textbf{34.3} & 56.0 & 36.4 & 15.4 & 38.1 & 51.9\\
\end{tabular}
\end{center}
\caption{Comparison of three region feature extraction methods using different support regions. Accuracies are reported on COCO detection \textit{minival} set. *It is not clear how to exploit the whole image for regular and aligned RoI pooling methods. Hence the corresponding accuracy numbers are omitted.}
\label{table.aggregation_range}
\end{table}

\subsection{Ablation Study}
\subsubsection{Effect of supporting region $\Omega$.} It is investigated in
Table~\ref{table.aggregation_range}. Three sizes of the supporting region $\Omega$ are compared: the RoI itself, the RoI expanded with twice the area (with the same center), and the whole image range. Regular and aligned RoI pooling are also compared\footnote{Deformable RoI pooling~\cite{dai2017deformable} is omitted as it does not have a fixed support region.}.

There are two observations. First, our method outperforms the other two pooling methods. Second, our method steadily improves from using larger support regions, indicating that exploiting contextual information is helpful. Yet, using larger support regions, e.g., $2\times$ RoI region, has minor and no improvements for regular and aligned RoI pooling, respectively, when compared to using $1\times$ RoI region. Moreover, it is unclear how to exploit the whole image for regular and aligned pooling in a reasonable way.

\subsubsection{Effect of sparse sampling.}
Table~\ref{table.aggregation_point_sampling} presents the results of using different numbers of sampling positions for efficient implementation. By utilizing proper number of sampling positions, the accuracy can be very close to that of naive dense enumeration. And the computational overhead can be significantly reduced thanks to the sparse sampling implementation. By default, 196 maximum sampling positions are specified for both $\Omega_b^{\text{In}}$ and $\Omega_b^{\text{Out}}$. The mAP score is 0.2 lower than that of dense enumeration. In runtime, large RoIs will have fewer sampling positions for $\Omega_b^{\text{Out}}$ and small RoIs will have fewer sampling positions than the maximum threshold for $\Omega_b^{\text{In}}$. The average counted sampling positions in runtime are are around 114 and 86 for $\Omega_b^{\text{In}}$ and $\Omega_b^{\text{Out}}$, respectively, as shown in Table~\ref{table.aggregation_point_sampling}. The corresponding computational cost is 4.16G FLOPS, which coarsely equals that of the 2-fc head (about 3.9G FLOPs).

For all the following experiments, our method will utilize the sparse sampling implementation with 196 maximum sampling positions for both $\Omega_b^{\text{In}}$ and $\Omega_b^{\text{Out}}$.

\renewcommand{\arraystretch}{1.1}
\begin{table}[t]
\small
\begin{center}
\begin{tabular}{cc|c|cc|ccc|ccc}
 \tabincell{c}{$|\Omega_b^{\text{Out}}|_{\text{max}}$} & $|\Omega_b^{\text{In}}|_{\text{max}}$ &  \tabincell{c}{\scriptsize mAP} &  \tabincell{c}{\scriptsize mAP$_{50}$} &  \tabincell{c}{\scriptsize mAP$_{75}$} & \tabincell{c}{\scriptsize mAP$_S$} & \tabincell{c}{\scriptsize mAP$_M$} & \tabincell{c}{\scriptsize mAP$_L$} & $|\Omega_b^{\text{Out}}|_\text{avg}$ & $|\Omega_b^{\text{In}}|_\text{avg}$  & FLOPS\\
\shline
\textit{full}* & $7^2$ & 33.4 & 55.6 & 35.3 & 14.1 & 37.2 & 50.7 & 1737 & 32 & 15.3G\\
\textit{full} & $14^2$ & 34.2 & 56.2 & 36.3 & 15.0 & 38.5 & 51.3 & 1737 & 86 & 15.7G\\
\textit{full} & $21^2$ & 34.1 & 56.0 & 35.9 & 14.5 & 38.3 & 51.1 & 1737 & 158 & 16.2G \\
\textit{full} & \textit{full} & 34.3 & 56.0 & 36.4 & 15.4 & 38.1 & 51.9 & 1737 & 282 & 17.1G\\
\hline
100 & $14^2$ & 33.8 & 55.5 & 35.9 & 14.3 & 38.0 & 50.8 & 71 & 86& 3.84G \\
\textbf{196} & $\mathbf{14^2}$ & 34.1 & 55.6 & 36.3 & 14.5 & 38.3 & 51.1 & \textbf{114} & \textbf{86} & \textbf{4.16G} \\
400 & $14^2$ &  34.0 & 55.7 & 36.0 & 14.4 & 38.4 & 51.0 & 194 & 86 & 4.72G \\
625 & $14^2$ & 34.1 & 55.7 & 36.1 & 14.5 & 38.0 & 51.3 & 432 & 86& 6.42G\\
\textit{full} & $14^2$ & 34.2 & 56.2 & 36.3 & 15.0 & 38.5 & 51.3 & 1737 & 86 & 15.7G\\
\end{tabular}
\end{center}
\caption{Detection accuracy and computational times of \textit{efficient} method using different number of sample points. The average samples $|\Omega_b^{\text{Out}}|_\text{avg}$ and $|\Omega_b^{\text{In}}|_\text{avg}$ are counted on COCO \textit{minival} set using 300 ResNet-50 RPN proposals. The bold row ($|\Omega_b^{\text{Out}}|_{\text{max}}=196$, $|\Omega_b^{\text{In}}|_{\text{max}}=14^2$) are used as our default maximum sample point number. *\textit{full} indicates that all image positions are used without any sampling.}
\label{table.aggregation_point_sampling}
\end{table}

\renewcommand{\arraystretch}{1.1}
\begin{table}[t]
\small
\begin{center}
\begin{tabular}{l|c|cc|ccc}
 \tabincell{c}{method} & \tabincell{c}{\scriptsize mAP} &
\scriptsize mAP$_{50}$ & \scriptsize mAP$_{75}$ & \tabincell{c}{\scriptsize mAP$_S$} & \tabincell{c}{\scriptsize mAP$_M$} & \tabincell{c}{\scriptsize mAP$_L$} \\
\shline
regular RoI pooling & 29.8 & 52.2 & 29.9 & 10.4 & 32.6 & 47.8\\
aligned RoI pooling & 32.9 & 54.0 & 34.9 & 13.9 & 36.9 & 48.8 \\
deformable pooling & 34.0 & 55.3 & 36.0 & 14.7 & 38.3 & 50.4 \\
\hline
our (geometry) & \textbf{33.2} & 55.2 & 35.4 & 14.2 & 37.0 & 50.0 \\
our (geometry+appearance) & \textbf{34.1} & 55.6 & 36.3 & 14.5 & 38.3 & 51.1\\
\end{tabular}
\end{center}
\caption{Effect of geometric and appearance terms in Eq.~(\ref{eq.att_weight}) for the proposed region feature extraction module. Detection accuracies are reported on COCO \textit{minival} set.}
\label{table.attentional_weight}
\end{table}

\subsubsection{Effect of geometric relation and appearance feature terms.}
Table~\ref{table.attentional_weight} studies the effect of geometric relation and appearance feature terms in Eq.~(\ref{eq.att_weight}) of the proposed module. Using geometric relation alone, the proposed module is slightly better than aligned RoI pooling, and is noticeably better than regular RoI pooling. By further incorporating the appearance feature term, the mAP score rises by 0.9 to 34.1. The accuracy is on par with deformable RoI pooling, which also exploits appearance features to guide the region feature extraction process.

\subsubsection{Comparison on stronger detection backbones.}
We further compare the proposed module with regular, aligned and deformable versions of RoI pooling on stronger detection backbones, where FPN and ResNet-101 are also utilized.

Table~\ref{table.backbones} presents the results on COCO \textit{test-dev} set. Using the stronger detection backbones, the proposed module also achieves on par accuracy with deformable RoI pooling, which is noticeably better than aligned and regular versions of RoI pooling. We achieve a final mAP score of 39.9 using FPN+ResNet-101 by the proposed fully learnable region feature extraction module.

\renewcommand{\arraystretch}{1.1}
\begin{table}[t]
\small
\begin{center}
\begin{tabular}{l|l|c|cc|ccc}
 \tabincell{c}{backbone} & method & \tabincell{c}{\scriptsize mAP} &
\scriptsize mAP$_{50}$ & \scriptsize mAP$_{75}$ & \tabincell{c}{\scriptsize mAP$_S$} & \tabincell{c}{\scriptsize mAP$_M$} & \tabincell{c}{\scriptsize mAP$_L$} \\
\shline
\multirow{2}{*}{\tabincell{c}{Faster R-CNN \\+ResNet-50}} & regular RoI pooling & 29.9 & 52.6 & 30.1 & 9.7 & 31.9 & 46.3 \\
 & aligned RoI pooling & 33.1 & 54.5 & 35.1 & 13.9 & 36.0 & 47.4\\
 & deformable RoI pooling & 34.2 & 55.7 & 36.7 & 14.5 & 37.4 & 48.8\\
 \cline{2-8}
 & our & \textbf{34.5} & 56.4 & 36.4 & 14.6 & 37.4 & 50.3\\
 \hline
 \multirow{2}{*}{\tabincell{c}{Faster R-CNN \\+ResNet-101}} & regular RoI pooling & 32.7 & 53.6 & 23.7 & 11.4 & 35.2 & 50.0 \\
 & aligned RoI pooling & 35.6 & 57.1 & 38.0 & 15.3 & 39.3 & 51.0\\
 & deformable RoI pooling & \textbf{36.4} & 58.1 & 39.3 & 15.7 & 40.2 & 52.1\\
  \cline{2-8}
 & our & \textbf{36.4} & 58.6 & 38.6 & 15.3 & 40.2 & 52.2\\
  \hline
 \multirow{2}{*}{\tabincell{c}{FPN \\+ResNet-50}} & regular RoI pooling & 35.9 & 59.0 & 38.4 & 19.6 & 38.8 & 45.4 \\
 & aligned RoI pooling & 36.7 & 59.1 & 39.4 & 20.9 & 39.5 & 46.3\\
 & deformable RoI pooling & 37.7 & 60.6 & 40.9 & 21.3 & 40.7 & 47.4\\
  \cline{2-8}
 & our & \textbf{37.8} & 60.9 & 40.7 & 21.3 & 40.4 & 48.0\\
  \hline
 \multirow{2}{*}{\tabincell{c}{FPN \\+ResNet-101}} & regular RoI pooling & 38.5 & 61.5 & 41.8 & 21.4 & 42.0 & 49.2 \\
 & aligned RoI pooling & 39.1 & 61.4 & 42.3 & 21.5 & 42.5 & 50.2\\
 & deformable RoI pooling & \textbf{40.0} & 62.7 & 43.5 & 22.4 & 43.4 & 51.3\\
  \cline{2-8}
 & our & 39.9 & 63.1 & 43.1 & 22.2 & 43.4 & 51.6\\
\end{tabular}
\end{center}
\caption{Comparison of different algorithms using different backbones. Accuracies on COCO \textit{test-dev} are reported.}
\label{table.backbones}
\end{table}

\begin{figure}
\begin{center}
\begin{tabular}{p{0.99\columnwidth}}
\includegraphics[width=0.99\textwidth]{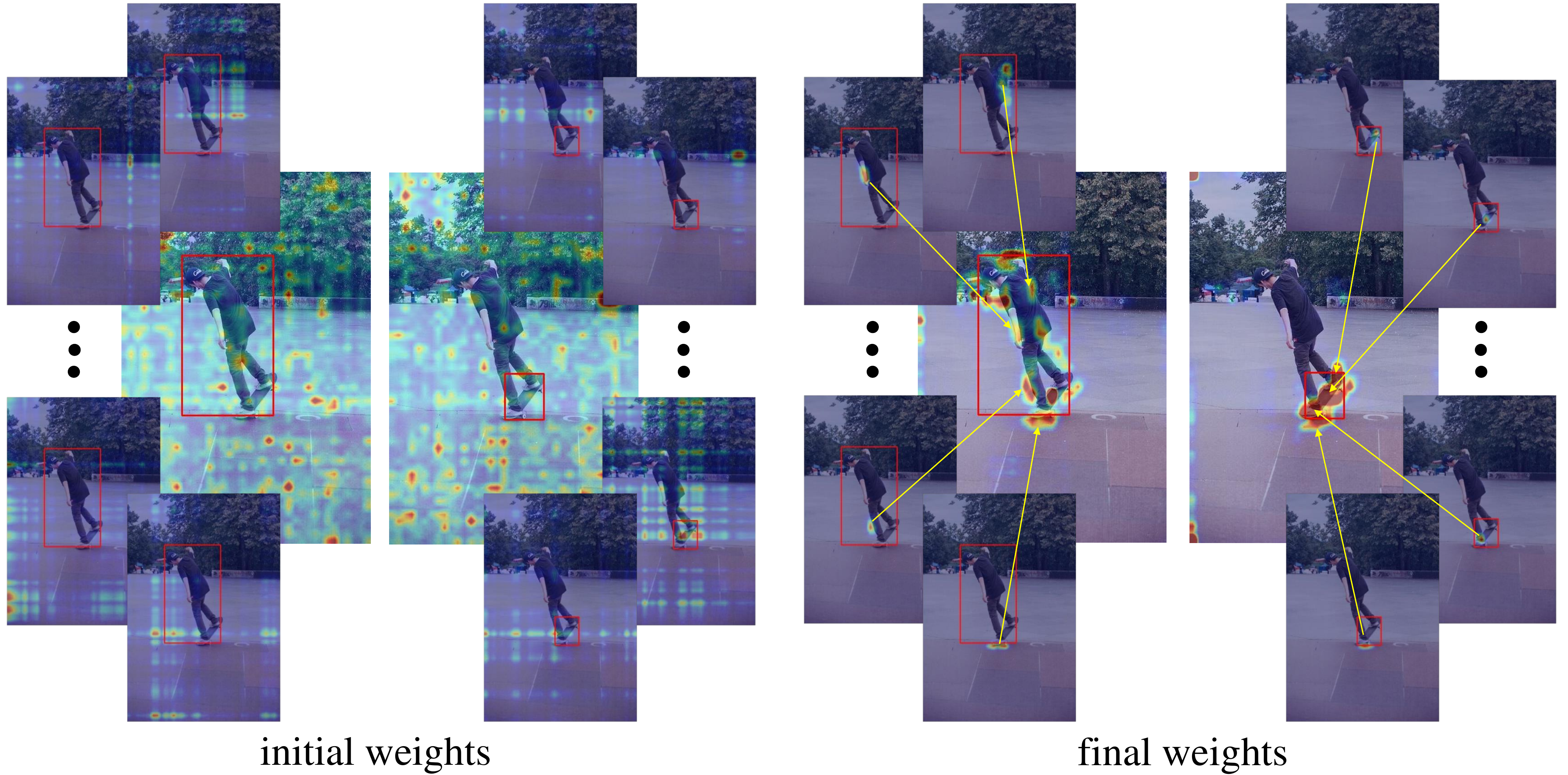} \\
{(a) The initial (\textbf{left}) and final (\textbf{right}) weights $w_k(*)$ in Eq.~(\ref{eq.att_weight}) of two given RoIs (the red boxes). The center images show the maximum value of all $K=49$ weight maps. The smaller images around show 4 individual weight maps. } \\
{}\\
\includegraphics[width=0.99\textwidth]{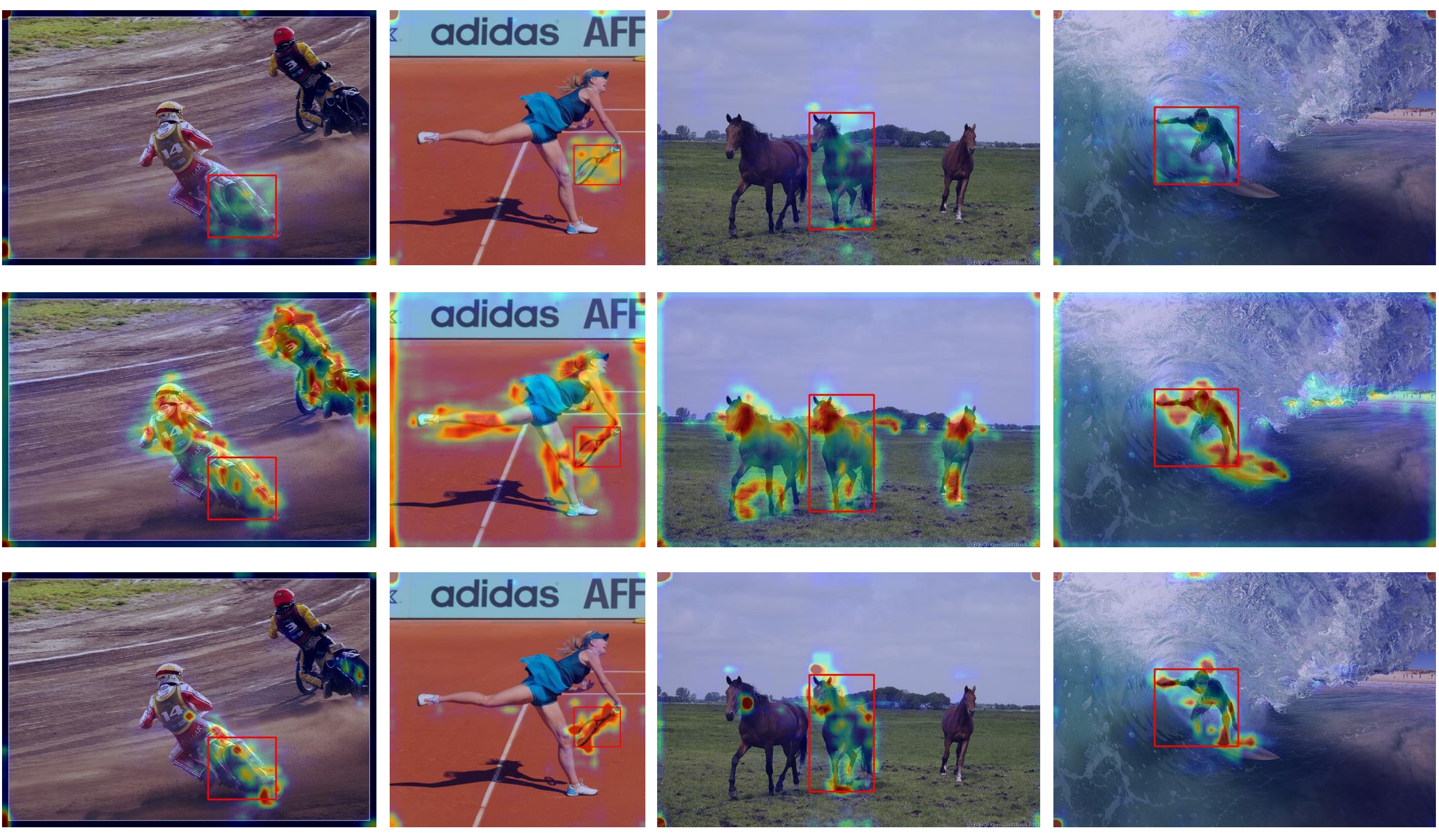}\\
{(b) Example results of geometric weights (\textbf{top}), appearance weights (\textbf{median}) and final weights (\textbf{bottom}).} \\
\end{tabular}
\end{center}
\caption{Qualitative analysis of learnt weights. For visualization, all weights are normalized by the maximum value over all image positions and half-half matted with the original image.}
\label{figure.att_weights}
\end{figure}

\section{What is learnt?}
\label{sec.learnt_weights}

\paragraph{Qualitative Analysis} The learnt weights $w_k(*)$ in Eq.~(\ref{eq.att_weight}) are visualized in Figure~\ref{figure.att_weights} (a). The supporting region $\Omega$ is the whole image.

Initially, the weights $w_k(*)$ are largely random on the whole image. After training, weights in different parts are learnt to focus on different areas on the RoI, and they mostly focus on the instance foreground.

To understand the role of the geometric and appearance terms in Eq.~(\ref{eq.att_weight}), Figure~\ref{figure.att_weights} (b) visualizes the weights when either of them is ignored. It seems that the geometric weights mainly attend to the RoI, while the appearance weight focuses on all instance foreground.

\paragraph{Quantitative Analysis} For each part $k$, the weights $w_k(*)$ are treated as a probability distribution over all the positions in the supporting region $\Omega$, as $\sum_{p \in \Omega}w_k(b,p,\mathbf{x}) = 1$. KL divergence is used to measure the discrepancy between such distributions.

\begin{figure}
\begin{center}
\begin{tabular}{p{0.99\columnwidth}}
\includegraphics[width=0.99\textwidth]{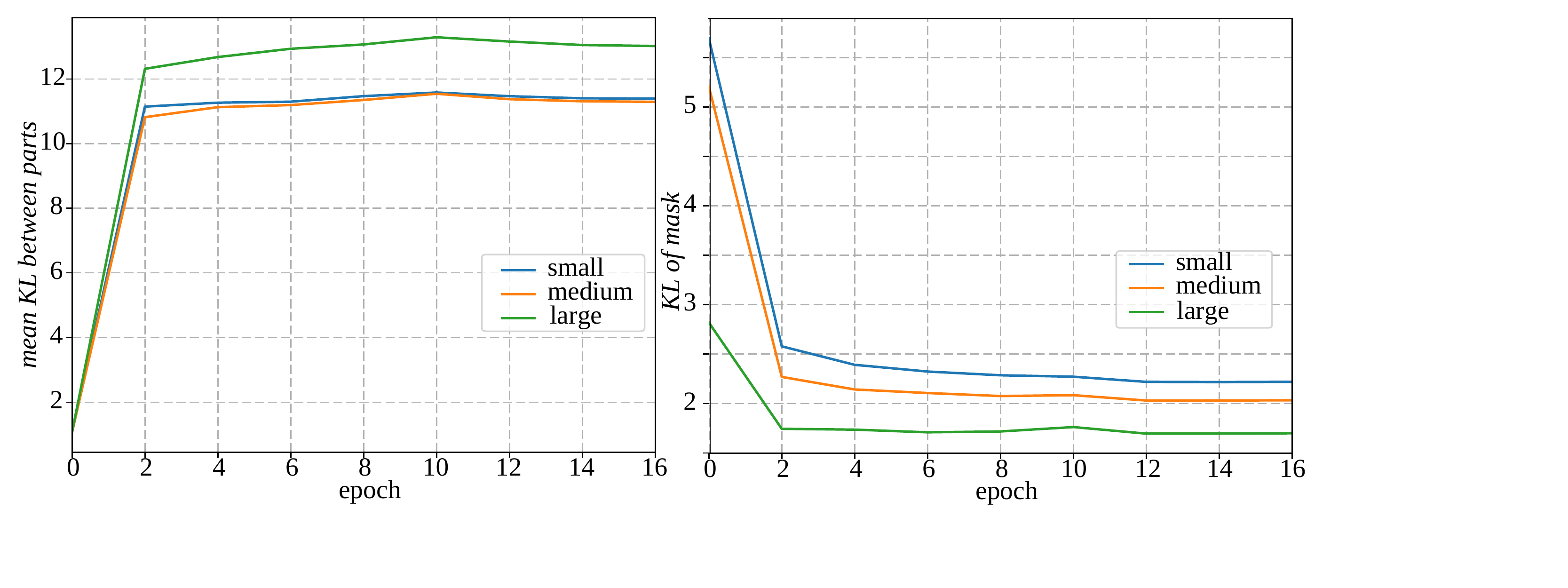}\\
\end{tabular}
\end{center}
\caption{Quantitative analysis of learnt weights. The two plots are \emph{mean KL between parts} (\textbf{left}) and \emph{KL of mask} (\textbf{right}) during training, respectively. Note that we test KL divergence every two epochs since our training framework saves model weights using such frequency.}
\label{figure.att_weights_quantitative}
\end{figure}

We firstly compare the weights in different parts. For each ground truth object RoI, KL divergence value is computed between all pairs of $w_{k_1}(*)$ and $w_{k_2}(*)$, $k_1,k_2=1,...,49$. Such values are then averaged, called \emph{mean KL between parts} for the RoI. Figure~\ref{figure.att_weights_quantitative} (left) shows its value averaged over objects of three sizes (as defined by COCO dataset) during training. Initially, the weights of different parts are largely indistinguishable. Their KL divergence measure is small. The measure grows dramatically after the first test. This indicates that \emph{the different parts are learnt to focus on different spatial positions}. Note that the divergence is larger for \emph{large} objects, which is reasonable.

We then investigate how the weights resemble the instance foreground, by comparing them to the ground-truth instance foreground mask in COCO. Towards this, for each ground truth object RoI, the weights from all the parts are aggregated together by taking the maximum value at each position, resulting in a ``max pooled weight map". The map is then normalized as a distribution (sum is 1). The ground truth object mask is filled with 1 and 0. It is also normalized as a distribution. KL divergence between these two distributions is called \emph{KL of mask}. Figure~\ref{figure.att_weights_quantitative} (right) shows this measure averaged over objects of three sizes during training. It quickly becomes small, indicating that \emph{the aggregation of all part weights is learnt to be similar as the object mask}.

The second observation is especially interesting, as it suggests that learning the weights as in Eq.~(\ref{eq.att_weight}) is related to instance segmentation, in some implicit manner. This is worth more investigation in the future work.

\bibliographystyle{splncs}
\bibliography{egbib}
\end{document}